\documentclass{aims}

\usepackage{amsfonts}
\usepackage{amsmath}
\usepackage{lmodern}
\usepackage[T1]{fontenc}
\usepackage{graphicx}
\usepackage{epstopdf}
\usepackage{algorithmic}
\usepackage{amssymb}
\usepackage{bm}
\usepackage{comment}
\usepackage{multirow}
\usepackage{hhline}
\usepackage{float}
\usepackage{caption}
\usepackage[T1]{fontenc}
\usepackage{tikz,caption}
\usepackage[caption=false,font=footnotesize]{subfig}
\usepackage{pgfplots}
\pgfplotsset{compat=newest}
\pgfplotsset{plot coordinates/math parser=false}
\usepackage{url}
\usepackage{color}
\usepackage{todonotes}

\usetikzlibrary{plotmarks} 
\usetikzlibrary{external}
 
\newlength\figureheight
\newlength\figurewidth 

\ifpdf
  \DeclareGraphicsExtensions{.eps,.pdf,.png,.jpg}
\else
  \DeclareGraphicsExtensions{.eps}
\fi

\graphicspath{ {pic/} }

\usepackage{xcolor}

\usepackage[pagewise]{lineno}\nolinenumbers
\usepackage{paralist}
\usepackage{graphics} 
\usepackage{epsfig} 
\usepackage{graphicx}  \usepackage{epstopdf} 
 \usepackage[colorlinks=true]{hyperref}
\hypersetup{urlcolor=blue, citecolor=red}

\def\currentvolume{X}
 \def\currentissue{X}
  \def\currentyear{200X}
   \def\currentmonth{XX}
    \def\ppages{X--XX}
     \def\DOI{10.3934/xx.xxxxxxx}

\theoremstyle{definition}

\newcommand{\rev}[1]{\textcolor{black}{#1}}

\newcommand{\R}{\mathbb{R}}
\def\norm#1{\left\|#1\right\|} 
\newcommand{\coloneqq}{\mathrel{\mathop:}=}

\newcommand{\ab}{{\bm{\alpha}}}

\newcommand{\data}[2]{\ensuremath{\bm{x}^{#2}_{#1}}}
\newcommand{\datauni}[2]{\ensuremath{x^{#2}_{#1}}}
\newcommand{\y}[1]{\ensuremath{y_{#1}}}
\newcommand{\anz}[0]{\ensuremath{N}}
\newcommand{\f}[0]{\ensuremath{d}}
\newcommand{\ffs}[0]{\ensuremath{d_\phi}}
\newcommand{\ord}[0]{\ensuremath{{d_\text{A}}}}

\newcommand{\IM}{{\mathcal I_{\bm M}}}
\newcommand{\x}{{\bm x}}
\newcommand{\Nat}{{\mathbb N}}
\newcommand{\C}{{\mathbb C}}
\newcommand{\T}{{\mathbb T}}

\newcommand{\e}{\mathrm e}
\newcommand{\ic}{\mathrm i}

\title[Learning Using ANOVA-Based Fast Matrix-Vector Multiplication] 
      {Learning in High-Dimensional Feature Spaces Using ANOVA-Based Fast Matrix-Vector Multiplication} 

\author[Franziska Nestler, Martin Stoll and Theresa Wagner]{}


 \email{franziska.nestler@math.tu-chemnitz.de}
 \email{martin.stoll@math.tu-chemnitz.de}
 \email{theresa.wagner@math.tu-chemnitz.de}

\thanks{$^*$Corresponding author: Theresa Wagner}

\begin{document}

\maketitle

\centerline{\scshape Franziska Nestler}
\medskip
{\footnotesize
 \centerline{Department of Mathematics, Chair of Applied Functional Analysis}
   \centerline{TU Chemnitz, Germany}
}

\medskip

\centerline{\scshape Martin Stoll and Theresa Wagner$^*$}
\medskip
{\footnotesize
  \centerline{Department of Mathematics, Chair of Scientific Computing}
   \centerline{TU Chemnitz, Germany}
} 

\bigskip

 \centerline{(Communicated by the associate editor name)}


\begin{abstract}
Kernel matrices are crucial in many learning tasks such as support vector machines or kernel ridge regression. The kernel matrix is typically dense and large-scale. Depending on the dimension of the feature space even the computation of all of its entries in reasonable time becomes a challenging task. For such dense matrices the cost of a matrix-vector product scales \rev{quadratically with the dimensionality $N$,} if no customized methods are applied. We propose the use of an ANOVA kernel, where we construct several kernels based on lower-dimensional feature spaces for which we provide fast algorithms realizing the matrix-vector products. We employ the non-equispaced fast Fourier transform (NFFT), which is of linear complexity for fixed accuracy. Based on a feature grouping approach, we then show how the fast matrix-vector products can be embedded into a learning method choosing kernel ridge regression and the conjugate gradient solver. We illustrate the performance of our approach on several data sets. 
\end{abstract}


\section{Introduction}
Kernel methods \cite{shawe2004kernel,hofmann2008kernel} are an important class of machine learning algorithms ranging from unsupervised to supervised learning. At the heart of these algorithms lies the kernel matrix $K\in\R^{N,N}$ with entries $K_{ij}:=\kappa \left( \x_i,\x_j \right)$, which arises from evaluating the kernel function $\kappa \left(\cdot,\cdot\right)$ on the given data points $\x_i\in\R^d$. The kernel matrix is typically dense and large-scale, which makes computations with it computationally demanding \cite{stoll2020literature,saad2003iterative,golub2013matrix}. In order to reduce the computational cost of working with dense kernel matrices many approaches have been considered in the literature. Among them fast transforms are a very promising group of methods. We here point to \cite{morariu2008automatic,raykar2007fast} for the fast Gauss transform and to \cite{alfke2018nfft,post02} for the use of the non-equispaced fast Fourier transform (NFFT), which we will focus on in this paper. \rev{For a comparison of the fast Gauss transform with the NFFT-approach we refer to \cite{alfke2018nfft}.} Alternatively, sketching has been considered as a technique for solving kernel ridge regression problems \cite{avron2017faster} both for the task of forming a suitable preconditioner and performing the matrix-vector products efficiently.
Furthermore, fast parallel solvers for kernel matrices~\cite{INV-ASKIT} have been proposed computing an approximate factorization of the kernel matrix in the fashion of $N$-body methods~\cite{ASKIT} with $\mathcal O(N\log N)$ work.

Fast transform-based methods are typically hit by a curse of dimensionality when the dimensionality $d$ of the feature vectors $\x_i=(x_i^1,\dots,x_i^d)^\top\in\R^d$ grows without any additional assumptions made on the kernel function. In order to carry over the power of the NFFT-based fast matrix-vector products, we here propose the use of an ANOVA kernel that is a combination of kernels relying on a small number of features and hence enables fast multiplication methods. While we focus on kernel ridge regression, the proposed technique can be embedded into a variety of kernel-based learning schemes such as deep kernel methods \cite{wilson2016deep}, Gaussian processes \cite{rasmussen2003gaussian} or graph-based learning \cite{kipf2016semi}.

\begin{figure}[htp]
\begin{center}
\includegraphics[scale=0.25]{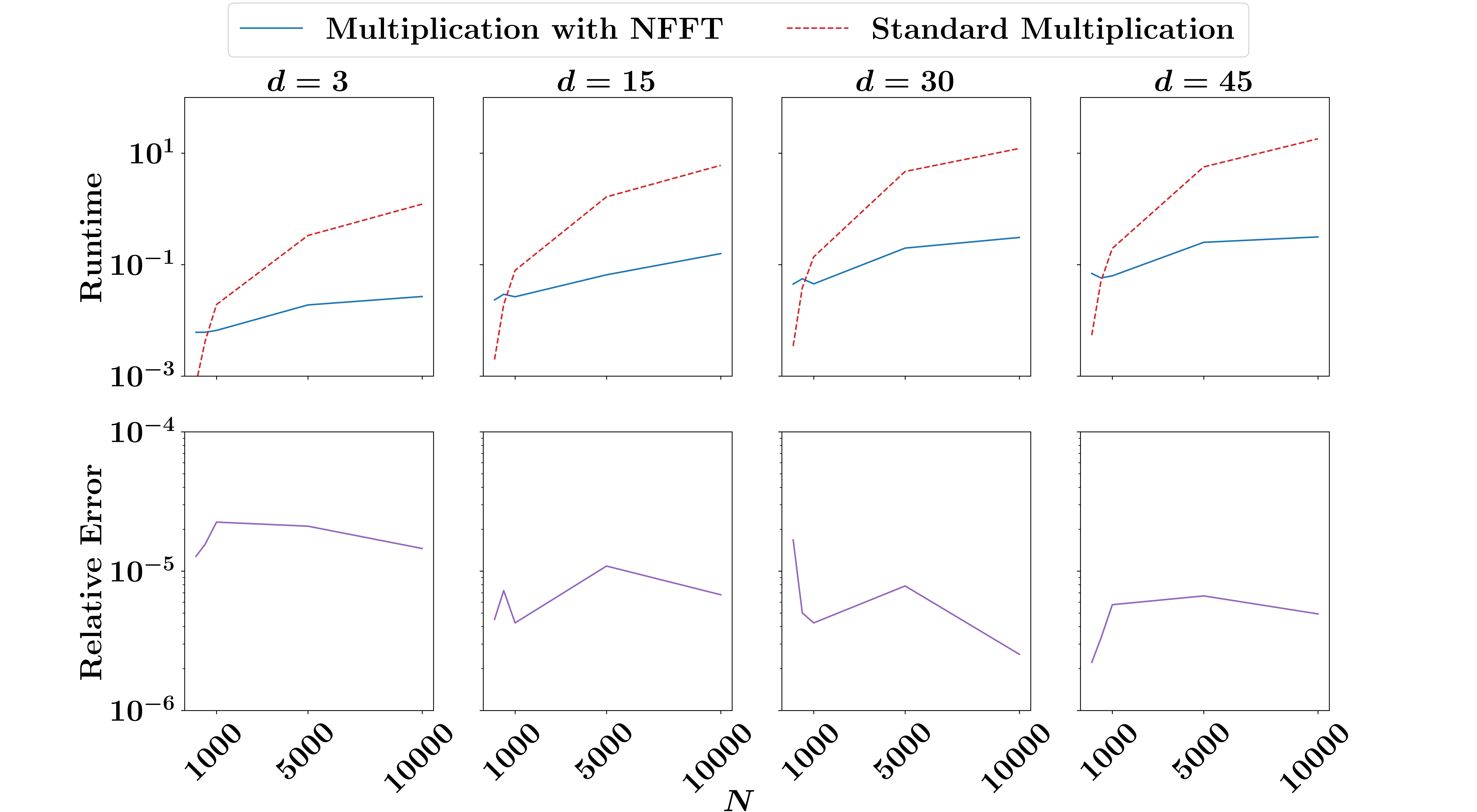}
\caption{Kernel-vector multiplication with and without NFFT-approach on various balanced samples of different size on synthetic data, with $\sigma=100$ and $n_{\text{runs}}=100$}
\label{fig:kernel_vec_mult_motivation}
\end{center}
\end{figure}

Figure~\ref{fig:kernel_vec_mult_motivation} indicates that the matrix-vector products using the NFFT will outperform the dense matrix-vector products even for small dimensions $N$. We use the remainder of this paper to discuss how this is achieved in detail. Naturally, the results will depend on the data, which for this example \rev{were generated randomly}, and the choice of parameters. Figure~\ref{fig:kernel_vec_mult_motivation} is meant to give an impression of the power of ANOVA-based fast matrix-vector multiplications.

We show in the remainder of this paper how such fast multiplications and a new feature grouping can be embedded into kernel-based learning. In Section~\ref{Kernel_Ridge_Regression} we introduce kernel ridge regression, which is followed by a discussion of the ANOVA kernel. We there show that this kernel allows a grouping of the features and after discussing the NFFT-based multiplication in Section~\ref{NFFT-based_Fast_Matrix_multiplication}, we introduce a novel multiple kernel grouping based on mutual information scores. Our experiments indicate that the new kernel results in much faster training and prediction times, especially for large data sets, without or with very benign loss of accuracy. These computations can be carried out on standard hardware without requiring additional high performance computing power.

\section{Kernel Ridge Regression}
\label{Kernel_Ridge_Regression}
As already eluded to, we focus on the case of kernel ridge regression as a learning method that utilizes positive kernels \cite{shawe2004kernel,hofmann2008kernel}. It is targeted at detecting dependencies between features and the label to successfully predict class affiliations for new data points from the same source. The underlying theory of machine learning is well developed for the linear case indeed \cite{shawe2004kernel}. Real-world problems typically require non-linear methods though. By learning in a reproducing kernel Hilbert space (RKHS)~\cite{scholkopf2002learning}, this difficulty can be overcome.
Replacing dot products by a kernel evaluation allows us to utilize linear methods in this space without ever explicitly having to compute in it (cf.\ \cite{ shawe2004kernel,hofmann2008kernel,alfke2018nfft}).

Let $\mathcal{H}$ be a RKHS
\begin{align*}
\rev{\mathcal{H}}:=\overline{\mathrm{span}\left\{\kappa(\x,\cdot), \x\in\mathcal X\right\}},
\end{align*}
where $\mathcal{X}$ is a non-empty set and $\kappa$ is a reproducing kernel of {$\mathcal{H}$} on {$\mathcal{X}\times \mathcal{X}$}. Moreover, let
\begin{align*}
    \langle f, g \rangle \coloneqq \sum_{j=1}^{\anz'} \sum_{i=1}^\anz \alpha_i \beta_j \kappa \left( {\data{j}{}}' , \data{i}{} \right)
\end{align*}
define the inner product of elements
\begin{align*}
    f \left( \cdot \right) = \sum_{i=1}^\anz \alpha_i \kappa \left( \cdot , \data{i}{} \right) \quad \text{and} \quad g \left( \cdot \right) = \sum_{j=1}^{\anz'} \beta_j \kappa \left( \cdot , {\data{j}{}}' \right)
\end{align*}
and
\begin{align*}
    \| f \|_{\mathcal{H}} \coloneqq \sqrt{\langle f , f \rangle}
\end{align*}
the norm in $\mathcal{H}$. Then, with $\kappa$ being the reproducing kernel of {$\mathcal{H}$}, the evaluation functional
\begin{align*}
    \langle \kappa \left( \cdot , \data{}{} \right) , f \rangle = \sum_{i=1}^\anz \alpha_i \kappa \left( \data{i}{} , \data{}{} \right) = f \left( \data{}{} \right)
\end{align*}
holds for all {$f \in \mathcal{H}$}, {$\data{}{} \in \mathcal{X}$}~\cite{scholkopf2002learning}.  Now, let a data set be given that is represented by feature vectors {$\data{i}{} \in \mathbb{R}^\f$}, {$i = 1, \dots, \anz$}, where {$\anz \in \mathbb{N}$} denotes the number of data points and {$\f \in \mathbb{N}$} the number of features. Furthermore, let each of those vectors possess a label {$\y{i} \in \{ -1, 1 \}$}, so that we obtain pairs {$\left( \data{i}{}, \y{i} \right)$}. As explained above, our aim is to train our model on the given labeled data, to be able to predict class affiliations for new unlabeled data later.

For that, we can formulate the learning task
\begin{align} \label{learning_problem_RKHS}
    \hat{f} = \underset{f \in \mathcal{H}}{\arg \text{min}} \sum_{i=1}^\anz \left( \y{i} - f \left( \data{i}{} \right) \right) ^2 + \lambda \| f \|_{\mathcal{H}}^2
\end{align}
in a RKHS $\mathcal{H}$, where $\lambda > 0$ is the regularization parameter \rev{(cf. \cite{bishop2006pattern})}. It corresponds to the linear regression problem in $\mathcal{X}$. By the representer theorem \cite{hofmann2008kernel}, the solution to~\eqref{learning_problem_RKHS} is of the form
\begin{align} \label{representer_theorem}
    f = \sum_{i=1}^\anz \alpha_i \kappa \left( \data{i}{}, \cdot \right),
\end{align}
where {$\data{i}{} \in \mathbb{R}^\f$}
are the labeled training samples~\cite{scholkopf2002learning}. \rev{Substituting}~\eqref{representer_theorem} to \eqref{learning_problem_RKHS}, we obtain the finite-dimensional quadratic program
\begin{align*}
    \hat{\ab} = \underset{\ab \in \mathbb{R}^\anz}{\arg \text{min}} \sum_{i=1}^\anz \left( y_i - \sum_{j=1}^\anz \alpha_j \kappa \left( \data{j}{}, \data{i}{} \right) \right) ^2 + \lambda \sum_{i=1}^\anz \sum_{j=1}^\anz \alpha_i \alpha_j \kappa \left( \data{i}{}, \data{j}{} \right)
\end{align*}
in {$\ab = \begin{bmatrix} \alpha_1, \dots, \alpha_\anz \end{bmatrix}^\top$}, see \cite{hofmann2008kernel} for details. This can be rewritten as
\begin{align} \label{nfft_KRR}
    \hat{\ab} = \underset{\ab \in \mathbb{R}^\anz}{\arg \text{min}} \| \bm{\y{}} - K \ab \|_2^2 + \lambda \ab^\top  K \ab,
\end{align}
where {$\bm{\y{}} = \begin{bmatrix} \y{1}, \dots, \y{\anz} \end{bmatrix}^\top$} is the target vector and
\begin{align*}
    K = \begin{bmatrix}
    \kappa \left( \data{1}{}, \data{1}{} \right) & \dots & \kappa \left( \data{1}{}, \data{\anz}{} \right) \\
    \vdots & & \vdots \\
    \kappa \left( \data{\anz}{}, \data{1}{} \right) & \dots & \kappa \left( \data{\anz}{}, \data{\anz}{} \right)
    \end{bmatrix}
\end{align*}
the kernel matrix, being symmetric and positive semi-definite. 
In a straightforward way, we obtain the gradient of the objective function as
\begin{align*}
K \left( K+\lambda I \right) \ab-K \bm{\y{}}\stackrel{!}{=}\bm 0
\end{align*}
and we get for the optimal $\hat{\ab}$ that
\begin{align*} \label{linear_system_KRR}
    \left( K + \lambda I \right)\hat{\ab}- \bm{\y{}}\in\ker(K),
\end{align*}
which is true for 
\begin{align*}
\hat{\ab}=\left( K + \lambda I \right)^{-1}\bm{\y{}}+\bm{z},
\end{align*}
where $K\bm{z}=0$. Due to the fact that for $\ab'=\ab+\bm{z}$ \rev{and $f' = \sum_{i=1}^\anz \alpha_i' \kappa \left( \data{i}{}, \cdot \right)$} 
\begin{align*}
    \norm{f-f'}^2_{\mathcal H}=(\ab-\ab')^\top K(\ab-\ab')=0,
\end{align*}
we have that $f=f'$ and the only solution we care for is obtained as
\begin{align*}
\hat{\ab}=\left( K + \lambda I \right)^{-1}\bm{\y{}}.
\end{align*}
For $\lambda>0$ the system to be solved is symmetric and positive definite. Hence, we will rely on the conjugate gradient (CG) method \cite{hestenes1952methods} for its solution. The CG method requires the computation of one matrix-vector product per iteration and, thus, the multiplication with the kernel matrix $K$
\begin{equation}\label{eq:mult_with_K}
K\boldsymbol\ab=\left[\sum_{j=1}^N \alpha_j\kappa\left(\x_i,\x_j\right)\right]_{i=1}^N \in\R^N
\end{equation}
is the most expensive task within this procedure. A naive direct computation requires $\mathcal O(N^2)$ arithmetic operations, which we will reduce drastically by applying a fast summation approach, as introduced in Section~\ref{sec:fastsum}.

\section{The ANOVA Kernel} \label{Kernel_Basics}
\begin{figure}[h]
\centering
\includegraphics[scale=0.325]{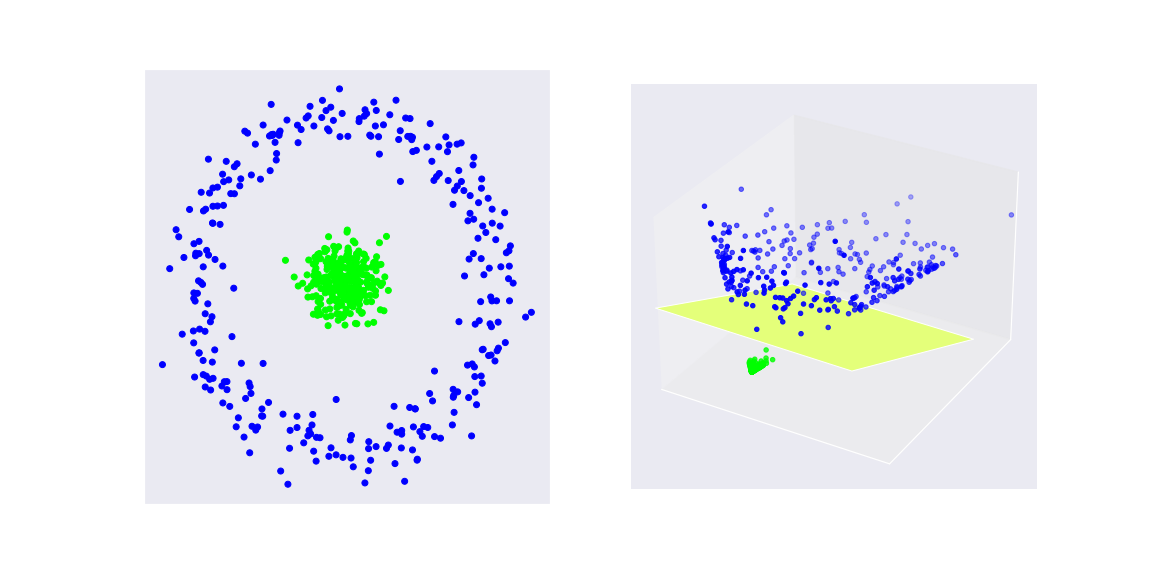}
\caption{Embedding 2-dimensional data into a 3-dimensional space via an embedding map.}
\label{fig:embedding_feature_space}
\end{figure}

As indicated above, kernel functions are derived from the use of inner products when the similarity between data points needs to be evaluated. Thereby, the data is embedded into a new, higher-dimensional feature space such that non-linear relations between features can be modeled in a linear way as illustrated in Figure~\ref{fig:embedding_feature_space}, see \cite{shawe2004kernel} for details. In this higher-dimensional space the separation of the data is easier to perform, which again can bee seen from the illustration in Figure~\ref{fig:embedding_feature_space}.

But the right choice of a kernel is by no means trivial. In this section we design a favorable kernel function for speeding up matrix-vector multiplications for large-scale data.

Let a data set be represented by feature vectors \rev{{$\x_i = (x_i^1,\dots,x_i^d)^\top \in\R^d$}}, where $\f$ is the number of features and $\anz$ is the number of data points or samples. After introducing an embedding map
\begin{align*}
    \phi: \data{}{} \in \mathbb{R}^\f \longmapsto \phi \left( \data{}{} \right) \in \mathbb{R}^{\ffs},
\end{align*}
with {$\f < \ffs$}, we recode our pairs of data points from {$\left( \data{i}{},\y{i} \right)$} to {$\left( \phi \left( \data{i}{} \right), \y{i} \right)$}. This enables us to transform non-linear relations to linear ones. Substituting all inner products {${\data{i}{}}^\top \data{j}{}$} between two feature vectors {$\data{i}{}, \data{j}{} \in \mathbb{R}^\f$} with the kernel evaluation
\begin{align*} 
    \kappa \left( \data{i}{}, \data{j}{} \right) = \langle \phi \left( \data{i}{} \right), \phi \left( \data{j}{} \right) \rangle
\end{align*}
is referred to as the \emph{kernel trick}~\cite{shawe2004kernel}.

The kernel crucial for our investigations in this paper is the well-known ANOVA kernel~\cite{shawe2004kernel}
\begin{align*} 
    \rev{\kappa_{\text{\tiny{anova}}}^\ord} \left( \data{i}{}, \data{j}{} \right) = \sum_{1 \leq l_1 < l_2 < \dots < l_\ord \leq \f} \prod_{t=1}^\ord \kappa_{l_t}^{\text{base}} \left( \data{i}{}, \data{j}{} \right)
\end{align*}
of degree $\ord$, with base kernels $\kappa_{l_t}^{\text{base}} \left( \data{i}{}, \data{j}{} \right)$ extending the kernel evaluation.

In comparison to other kernels, it provides more freedom regarding the choice of considered features and the weighting (cf.~\cite{shawe2004kernel}).

Given the importance of the Gaussian kernel~\cite{rasmussen2003gaussian}
\begin{align*}
    \rev{\kappa_{\text{\tiny{gauss}}}} \left( \data{i}{}, \data{j}{} \right) = \exp \left( - \frac{\| \data{i}{} - \data{j}{} \|^2_2}{\sigma^2} \right),
\end{align*}
we use weighted Gaussian kernels for all our base kernels. Moreover, if we set the degree $\ord = 1$, this yields the so-called \emph{simple Gaussian ANOVA kernel}
\begin{align*}
    \rev{\kappa_{\text{\tiny{anova}}}^{\text{\tiny{1d-gauss}}}} \left( \data{i}{}, \data{j}{} \right) = \sum_{l=1}^{\f} \eta_l \exp \left( - \frac{\| x_i^l - x_j^l \|_2^2}{\sigma^2} \right),
\end{align*}
in which the $l$-th Gaussian base kernel only depends on the $l$-th feature. This yields a sum of $\f$ weighted kernels. Each feature is taken into account in this kernel definition, which seems reasonable. But including particular features, which are either redundant or do not interact with the label at all, attributes more influence to those features than they actually have. By this, relations might be detected, which do not exist in reality. Excluding these detrimental features instead, can help us increase the model's accuracy. Examining each feature's importance separately instead neglects relevant relations between them. Additionally, we must not forget that we aim to minimize the computational complexity. We therefore are well-advised to keep the number of summands low. This gives rise to the \emph{extended Gaussian ANOVA kernel}
\begin{align} \label{extended_Gaussian_ANOVA_kernel}
    \rev{\kappa_{\text{\tiny{anova}}}^{\text{\tiny{3d-gauss}}}} \left( \data{i}{}, \data{j}{} \right) = \sum_{l=1}^P \eta_l \underbrace{\exp \left( - \frac{\| \data{i}{\mathcal W_l} - \data{j}{\mathcal W_l} \|_2^2}{\sigma^2} \right)}_{\rev{\kappa_l \left( \data{i}{\mathcal{W}_l}, \data{j}{\mathcal{W}_l} \right)}},
\end{align}
where {$P \leq \lceil \frac{\f}{3} \rceil$}, with index sets
\begin{equation}\label{eq:index_window}
\rev{\mathcal W_l = \{ w_1^l, w_2^l, w_3^l\} \subseteq \{1, \dots, \f\}}
\end{equation}
and corresponding
\begin{equation}\label{eq:data_window}
\data{i}{\mathcal W_l} = \begin{bmatrix} \datauni{i}{w_1^l} & \datauni{i}{w_2^l} & \datauni{i}{w_3^l} \end{bmatrix}^\top
\quad \text{and} \quad
\data{j}{\mathcal W_l} = \begin{bmatrix} \datauni{j}{w_1^l} & \datauni{j}{w_2^l} & \datauni{j}{w_3^l} \end{bmatrix}^\top,
\end{equation}
where {$i,j = 1, \dots, \anz$}.
By this, exactly $3$ features embody the input for every Gaussian base kernel in \eqref{extended_Gaussian_ANOVA_kernel}. The motivation for using $3$-dimensional inputs is down to the NFFT, see Section~\ref{NFFT-based_Fast_Matrix_multiplication}, which provides the underlying mathematical theory for the presented method for speeding up matrix-vector multiplications for large-scale data. In fact, this method runs our computations very efficiently as long as the input dimension is smaller than $4$. \rev{Thus, $\mathcal{W}_l$ holding exactly $3$ features enables us to exploit the full computational power of the NFFT, while involving as many feature interactions as possible and keeping the number of summands low at the same time. This choice hence fulfills all crucial aspects. Also we choose the indices corresponding to each $\mathcal{W}_l$ only once so that the kernels never share any features.}
We refer to Section~\ref{Multiple_Kernel_Learning} for details on determining the index sets $\mathcal W_l$ and the kernel weights $\eta_l$.

\section{NFFT-Based Fast Matrix Multiplication} \label{NFFT-based_Fast_Matrix_multiplication}
In this section, we give a short introduction to the NFFT as well as the fast summation approach. With these methods, we are able to reduce the arithmetic complexity from $\mathcal O(N^2)$ to $\mathcal O(N)$.

\subsection{NFFT}
For an even grid size $\bm M=\left(M_1,\dots,M_d\right)^\top\in2\Nat^d$, i.\,e.\ a $d$-dimensional vector with even integer components, we consider multivariate index sets of the form
\begin{align*}
\IM:=\left\{-\tfrac{M_1}{2},\dots,\tfrac{M_1}{2}-1\right\}
\times\dots\times
\left\{-\tfrac{M_d}{2},\dots,\tfrac{M_d}{2}-1\right\}.
\end{align*}
For a given set of Fourier coefficients $\hat f_{\bm k} \in\C$, $\bm k \in\IM$, on that grid and a set of nodes $\x_j\in\T^d$, $j=1,\dots,N$, where the torus $\T$ is defined as $\T=\mathbb{R}\,\mathrm{mod}\,\mathbb{Z} \simeq [-\frac12,\frac12)$, the NFFT~\cite{duro93,bey95,st97,DuSc} is an efficient algorithm computing the sums
\begin{align}\label{eq:nfft}
f_j:=f\left(\x_j\right):=\sum_{\bm k \in \IM} \hat f_{\bm k} \e^{2\pi\ic\bm{k}^\top\x_j}\quad \forall j=1,\dots, N,
\end{align}
i.\,e., the function values of a 1-periodic trigonometric polynomial $f=f\left(\x\right)$ at arbitrary, commonly irregularly distributed nodes $\x_j$.
The arithmetic complexity is $\mathcal O\left(|\IM|\log|\IM|+N\right)$.
The accuracy is controlled by several additional parameters which we do not introduce here.
For a detailed discussion on parameters involved in applied algorithms we refer to~\cite{KeKuPo09}.
Note that the size of the grid $\IM$ grows exponentially with $d$, since $\min\left(\bm M\right)^d\leq|\IM|$.
In other words, the curse of dimensionality makes the evaluation expensive in case of large dimensions.
In order to overcome this problem, we exclusively deal with small dimensions $\leq 3$ in \rev{every NFFT computation}, where FFT-based methods on regular grids are still very efficient.
Thus, the \rev{overall feature dimension}, which we expect to be large, \rev{is decomposed into multiple ANOVA kernels of degree $\ord \leq 3$}, see Section~\ref{Multiple_Kernel_Learning} \rev{for the multiple kernel approach}.

A fast algorithm for the adjoint problem, which performs the summation
\begin{equation}\label{eq:adjnfft}
g_{\bm k}:= \sum_{j=1}^N c_j \e^{-2\pi\ic\bm{k}^\top\x_j} \quad \forall \bm{k} \in\IM
\end{equation}
efficiently, is known as adjoint NFFT or sometimes type-2 NFFT. Here, the nodes $\x_j\in\T^d$ and corresponding coefficients $c_j\in\mathbb C$ are given.
Using matrix-vector notation, equations \eqref{eq:nfft} and \eqref{eq:adjnfft} can be written as
\begin{align*}
\bm f:=\Phi\hat{\bm f} \quad\text{ and }\quad \bm g:=\Phi^*\bm c,
\end{align*}
respectively, where $\Phi=\left(\e^{2\pi\ic\bm k^\top\x_j}\right)_{j,\bm k}\in\mathbb C^{N,|\IM|}$ is a non-uniform Fourier matrix and $\Phi^*$ is simply its adjoint.
The NFFT and its adjoint are the main ingredients of the fast summation approach, which is introduced below.

\subsection{Fast Summation\label{sec:fastsum}}
The fast summation approach computes sums of the form
\begin{equation}\label{eq:kernelsum}
s\left(\bm z_i\right):=\sum_{j=1}^{N_x} \alpha_j \kappa\left(\bm z_i,\x_j\right) \quad\forall i=1,\dots,N_z,
\end{equation}
for two given sets of nodes $\mathcal Z:=\{\bm z_i\in\mathbb R^d,i=1,\dots,N_z\}$, $\mathcal X:=\{\bm x_j\in\mathbb R^d, j=1,\dots,N_x\}$, coefficients $\{\alpha_j\in\mathbb C,j=1,\dots,N_x\}$ \rev{and a given kernel $\kappa$} efficiently based on combining the above described algorithms (NFFT and adjoint NFFT), cf.~\cite{post02}.
Note that the special case $\mathcal X=\mathcal Z$ is included.
In the context of kernel ridge regression, such sums have to be computed when evaluating the model~\eqref{representer_theorem} at new data points, i.\,e.\ $\mathcal X$ represents the set of training data and $\mathcal Z$ contains new test data.
Within the kernel ridge regression we make use of the special case $\mathcal Z=\mathcal X$ with $N_x=N_z=:N$ in order to multiply with the kernel matrix $K$, see \eqref{eq:mult_with_K}.

\rev{While in our setting the kernel $\kappa$ in~\eqref{eq:kernelsum} represents a single summand $\kappa_l$ of our specially designed extended Gaussian ANOVA kernel~\eqref{extended_Gaussian_ANOVA_kernel}, the following discussion is also true for more general kernels.}

Note that we restrict our considerations to \rev{radial} kernels without singularities.
In a more general setting, also kernels with a singularity in the origin, as for example $\|\bm z-\x\|^{-1}$ or $\log\|\bm z-\x\|$ can be handled \cite{FeSt03,postni04}.
In this case we require $\bm z_i\neq \x_j$ for all $i,j$ or rather to exclude all terms with $i=j$ in the case that the two sets of nodes coincide.

The basic idea of the fast summation is explained as follows.
In the one-dimensional setting we approximate \rev{a} univariate kernel by a trigonometric polynomial
\begin{equation}\label{eq:kernelapprox}
\kappa\left(z,x \right)=\kappa\left( r \right)\approx \sum_{k=-M/2}^{M/2-1} \hat c_k\e^{2\pi\ic k \rev{r}/h},
\end{equation}
where $r:=z-x$ and the period $h$ has to be chosen appropriately.
Note that we consider non-periodic functions or rather kernels $\kappa$ and, thus, an approximation by a trigonometric polynomial is not straightforward.
Instead of truncating the function and simply sticking together periodic images, which would give a non-smooth periodic function, we introduce a larger period $h$ and construct a smooth transition in order to end up with a periodic function of some higher smoothness.

\begin{figure}[htp]
\centering
\begin{tikzpicture}
\begin{axis}[
width=11cm, height=5cm,
axis x line=none,
axis y line=none,
xmin=-2,xmax=4.5,
ymin=-0.3,ymax=1.1]

\addplot[color=black,line width=0.4pt] coordinates {(-1.5,0) (4,0)};
\addplot[smooth,line width=0.4pt] table[x=x,y expr=\thisrow{y}] {pic/fastsum_table.txt};

\addplot[line width=0.3pt,mark=|] coordinates {(-0.75,0)} node[below] {\small$-L$};
\addplot[line width=0.3pt,mark=|] coordinates {(0.75,0)} node[below] {\small$L$};
\addplot[line width=0.3pt,mark=|] coordinates {(-1.25,0)} node[below] {$-\frac h2$};
\addplot[line width=0.3pt,mark=|] coordinates {(1.25,0)} node[below] {$\frac h2$};
\addplot[line width=0.3pt,mark=|] coordinates {(3.75,0)} node[below] {$\frac{3h}2$};

\addplot[line width=0.3pt,mark=|] coordinates {(1.75,0)} node[below] {\small$L+\ell$};

\addplot[line width=0.3pt,mark=*] coordinates {(-0.75,0.2821)};
\addplot[line width=0.3pt,mark=*] coordinates {(0.75,0.2821)};
\addplot[line width=0.3pt,mark=*] coordinates {(1.75,0.2821)};
\addplot[line width=0.3pt,mark=*] coordinates {(3.25,0.2821)};

\node (A) at (axis cs:1.15,0.5){};
\node (B) at (axis cs:0.75,0.3){};
\draw[->] (A)--(B);

\node (A) at (axis cs:1.35,0.5){};
\node (B) at (axis cs:1.75,0.3){};
\draw[->] (A)--(B);

\addplot[color=black,mark=text,/pgf/text mark={\small smooth}] coordinates{(1.3,0.53)};
\addplot[color=black,mark=text,/pgf/text mark={$p\left(r\right)$}] coordinates{(1.3,0.2)};
\addplot[color=black,mark=text,/pgf/text mark={$\kappa \left(r\right)$}] coordinates{(0,0.53)};

\end{axis}
\end{tikzpicture}
\caption{The \rev{univariate} kernel function $\kappa$, here a Gaussian, is truncated and a polynomial $p$ is constructed such that we end up with a periodic function $\tilde\kappa$, which is smooth up to a certain degree.
In higher-dimensional settings, the kernel function is truncated at $\|\bm r\|=L$, i.\,e., radially.
The resulting periodic function will then have the period $h$ with respect to each dimension.
\label{fig:fastsum}}
\end{figure}
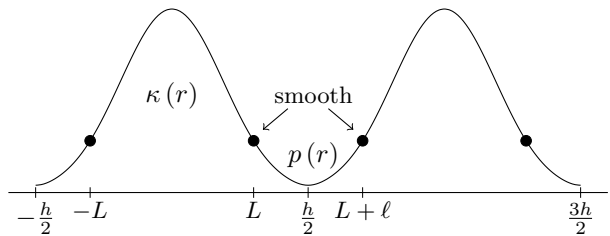

If the distances between the given nodes satisfy $|z_i-x_j|\leq L$, we choose $h\geq 2L$.
In order to get a good approximation we finally embed $\kappa$ into a smooth periodic function via
\begin{align*}
\tilde\kappa\left(r\right)=
\begin{cases}
\kappa\left(r\right) &: r\in [-L,L] ,\\
p\left(r\right) &: r\in \left(L,L+\ell\right),
\end{cases}
\end{align*}
where $p$ is a polynomial matching the derivatives of $\kappa$ up to a certain degree such that its periodic continuation $\tilde\kappa$ is smooth \cite{FeSt03}.
For a graphical illustration see Figure~\ref{fig:fastsum}.
We make heavy use of the fact that the kernel is symmetric.
If the given nodes lie in an interval of length $L$, the resulting period $h$ is more than twice as large as the original interval length.
The Fourier coefficients $\hat c_k$ \rev{in \eqref{eq:kernelapprox}} are simply computed by applying the ordinary FFT to $M$ equidistant samples of the resulting periodic function.

With some slight modifications this ansatz is also applicable in a multivariate setting with a radial kernel $\kappa\left(\|\cdot\|\right)$.
In this case, the problem is, roughly speaking, at first considered in an appropriate one-dimensional setting and the multivariate approximation is obtained via rotation \cite{FeSt03}. 
Defining the scaled nodes ${\tilde{\bm z}_i:=h^{-1}\bm z_i\in\T^d}$ and {$\tilde{\bm x}_j:=h^{-1}\bm x_j\in\T^d$}, we obtain
\begin{equation} \label{eq:fastsum}
\begin{aligned}
s\left(\bm z_i\right)
&\approx \sum_{j=1}^{N_x} \alpha_j \sum_{\bm k \in\IM} \hat c_{\bm k} \e^{2\pi\ic\bm{k}^\top\left(\tilde{\bm z}_i-\tilde\x_j\right)} \\
&=\sum_{\bm k \in\IM} \hat c_{\bm k} \left(\sum_{j=1}^{N_x}\alpha_j \e^{-2\pi\ic\bm{k}^\top\tilde\x_j}\right) \e^{2\pi\ic \bm{k}^\top\tilde{\bm z}_i}
\end{aligned}
\end{equation}
by replacing the kernel $\kappa$ in \eqref{eq:kernelsum} by its Fourier representation~\eqref{eq:kernelapprox}. The inner sums are now obtained via the adjoint NFFT~\eqref{eq:adjnfft}, followed by a multiplication with the Fourier coefficients.
Finally, the outer sums are computed with the NFFT~\eqref{eq:nfft}.

The quality of the approximation depends on several parameters, involving the grid size $\bm M$.
If these parameters, and therewith the basic accuracy of the approximation \eqref{eq:kernelapprox}, are kept fixed, the number of required arithmetic operations scales like $\mathcal O(N_x+N_z)$.

\subsection{Fast Multiplication with the Kernel Matrices}
Now, it is easy to see that the multiplication with the kernel matrix $K$ can be realized with the presented fast summation approach, \rev{since in our setting we simply consider a sum of radial kernels \eqref{extended_Gaussian_ANOVA_kernel} acting in low dimensions}.
Here we are in the special setting that the set of nodes $\mathcal X$ and $\mathcal Z$ coincide, as it is the case within the kernel ridge regression.

In matrix-vector notation, we have the approximation
\begin{align*}
K\boldsymbol\ab\approx \Phi D\Phi^*\boldsymbol\ab,
\end{align*}
where $\Phi=\left(\e^{2\pi\ic\bm{k}^\top\tilde\x_j}\right)_{j,\bm{k}}$ is defined as before and $D=\mathrm{diag}\left(\hat c_{\bm k}\right)_{\bm k}$ is a diagonal matrix
containing the precomputed Fourier coefficients of the periodized kernel $\tilde\kappa$.

After the training procedure, the predicted values for the testing data are obtained based on \eqref{eq:fastsum}.
The matrix-vector notation reads as
\begin{align*}
\bm s\approx \Phi_{\bm z} D\Phi^*\boldsymbol\ab,
\end{align*}
where the Fourier matrix $\Phi_{\bm z}:=\left(\e^{2\pi\ic\bm{k}^\top\tilde{\bm z}_i}\right)_{i,\bm{k}}$ now includes the test nodes ${\bm z}_i$.

\section{Multiple Kernel Learning} \label{Multiple_Kernel_Learning}

The NFFT-approach from Section~\ref{NFFT-based_Fast_Matrix_multiplication} is very fast for input-dimensions up to 3. However, the presented method is targeted at large-scale problems with a large number of features. The ANOVA kernel, which was introduced in Section~\ref{Kernel_Basics}, helps us decompose the feature dimensions into multiple kernels. That enables us to exploit the speed-up of the NFFT-approach anyhow. Therefore, we are in need of combining multiple kernels. Thus, a strategy on how to identify important features and on how to separate them up into multiple kernels is needed. 
Given its low computational complexity, we decide to rank all features by their mutual information score (MIS)~\cite{battiti1994using}.

Mutual information is a measure of uncertainty, i.\,e.\ it describes how much knowing the value of a feature reduces our uncertainty regarding the label. The mutual information score is a positive value that usually does not exceed 2. Since it is a univariate metric, it cannot detect relations between multiple features. Therefore, we must keep in mind that having a high mutual information score does not directly indicate a feature's real importance. However, mutual information has proven to yield reasonable results and is very attractive regarding its computational complexity. Thus, features with a score below a chosen threshold are dropped, all others are distributed to the index sets ${\mathcal W_l}$, see \eqref{eq:index_window}, for the kernel functions ${\kappa_l}$ in \eqref{extended_Gaussian_ANOVA_kernel}. All but (possibly) the last kernel are built from exactly $3$ features. For simplicity, they are arranged following the MIS ranking, such that the $3$ features with the highest scores are included in the first kernel $K_1$ and so on.

In most applications, using multiple kernels is motivated in combining different similarity measures or including samples coming from various sources. A prominent example from computational biology is combining protein sequences, gene expressions and protein structures to predict protein-protein interactions~\cite{tanabe2008simple}. But associating kernels based on such heterogeneous sets of data is by no means trivial. Gönen et al.~\cite{gonen2011multiple} give an excellent overview of common combination methods.

However, in our case, working with multiple kernels is due to the model construction. Therefore, we are in a completely different setting. The kernels neither rely on different similarity measures, nor do our data points come from various sources. 
Using an additional weighting scheme is not necessary here. Because of this, all kernels are equally weighted, so that their weights sum up to $1$. This yields the combined kernel
\begin{align*}
    K = \eta_1 K_1 + \dots + \eta_P K_P,
\end{align*}
with $\eta_l = \tfrac{1}{P}$ and
\begin{align*}
    K_l = \begin{bmatrix}
    \kappa_l \left( \data{1}{\mathcal W_l}, \data{1}{\mathcal W_l} \right) & \dots & \kappa_l \left( \data{1}{\mathcal W_l}, \data{\anz}{\mathcal W_l} \right) \\
    \vdots & & \vdots \\
    \kappa_l \left( \data{\anz}{\mathcal W_l}, \data{1}{\mathcal W_l} \right) & \dots & \kappa_l \left( \data{\anz}{\mathcal W_l}, \data{\anz}{\mathcal W_l} \right)
    \end{bmatrix}
\end{align*}
for ${l = 1, \dots, P}$. Here, ${P \in \mathbb{N}}$ is the number of kernels to combine and $\mathcal W_l$ are the considered index sets as defined in~\eqref{eq:index_window}, so that $\data{i}{\mathcal W_l}$ and $\data{j}{\mathcal W_l}$ are the corresponding data points restricted to those indices, see~\eqref{eq:data_window}.
Thereby, as requested, all inputs ${\data{i}{\mathcal W_l}}$, ${i = 1, \dots, \anz}$, ${l = 1, \dots, P-1}$, are ${3}$-dimensional. The last index set ${\mathcal W_P}$ contains all features, which are left, i.\,e.\ either $1$, $2$ or $3$ features.

The multiplication with $K$ is then performed using $P$ NFFTs. \rev{Note that the applicability of the fast NFFT-approach is of course not limited to equal weights.}

\section{Numerical Experiments}
In this section, we demonstrate numerical results for the presented method on a couple of benchmark data sets. The corresponding Python implementations are available at \url{https://github.com/wagnertheresa/NFFT4ANOVA}. They rely on the FastAdjacency package\protect\footnotemark by Bünger, which was introduced in \cite{alfke2018nfft}. All experiments in this paper were run on a standard laptop computer with \emph{8 $\times$ AMD Ryzen 7 4700U processors with Radeon Graphics} and \emph{15.0 GiB of RAM}. I.\,e.\ the proposed method does not require expensive hardware or having access to supercomputers, what enables a wide and low-threshold application.

\footnotetext{\url{https://github.com/dominikbuenger/FastAdjacency}}

\subsection{Data Preprocessing}
Operations such as computing the inner product or the norm obviously cannot be performed on every data type. Therefore, the data must be preprocessed in such a way that data points are free of missing values and categorical variables and solely consist of real values, before applying the classifier. In our implementations, we assume the data to already be of the required form. Therefore, (possibly) necessary preprocessing routines must be run individually beforehand. 

For handling missing values, we suggest to replace missing entries by some valid number. For details on imputation, we refer to Donders et al.~\cite{donders2006gentle}. Categorical variables are usually incorporated by applying one of numerous encoding schemes, such as the bin-counting scheme or the feature hashing scheme~\cite{zheng2018feature}. 

While dealing with missing values and categorical variables is not covered in our implementations, handling imbalanced data sets is. Since the KRR problem~\eqref{nfft_KRR} is targeted at maximizing the accuracy and minimizing the prediction-error, running the model on an imbalanced data set can have unwanted effects. Running the algorithm naively on a data set with one class being extremely over-represented might lead to a model, which assigns all future data to this class, no matter what the particular data points look like. In this case, the model did not discover anything during learning on the training data and all new data points from the under-represented class will be misclassified. However, these are precisely the cases, where classifying samples from the under-represented class correctly, i.\,e.\ achieving a high recall, is of utmost importance. Fraud detection is just one example for this. To ensure that this scenario never occurs and the model yields reasonable results, we use resampling techniques~\cite{marques2013suitability} to balance the class distribution of the data. Our model is targeted at large-scale data. Hence, for simplicity, we under-sample the over-represented class by randomly removing samples from the majority class, when a data set is imbalanced. 

Another necessary aspect of data preprocessing is scaling~\cite{adeyemo2019effects, patro2015normalization}. To make sure that our model does not perceive features with large scales as more relevant than features with small ranges, our implementation includes a function that z-score normalizes~\cite{sarlecomp} the data set before starting any computations. It standardizes all columns, such that their mean is $0$ and their standard deviation is $1$. By this, the impact of outliers is reduced. To prevent train-test contamination, mean and standard deviation are computed solely with respect to the training data, so that the test data are scaled with the same statistics as the training data.

\subsection{Accuracy and Runtime of NFFT-Based KRR Learning}
\label{Accuracy_and_Runtime_of_NFFT-Based_KRR_Learning}

To demonstrate the power of the NFFT-based fast summation, we want to apply our method to kernel ridge regression learning now. As described in Chapter~\ref{Kernel_Ridge_Regression}, computing the kernel-vector products within in CG algorithm, see~\eqref{eq:mult_with_K}, is most expensive within the learning task. We therefore want to replace the standard multiplication by the multiplication with NFFT-approach there to reduce the computational complexity.

For this, we consider a couple of benchmark data sets, see Table~\ref{table:benchmark_data_sets}.

\begin{table}[htp]
\begin{center}
\begin{tabular}{|c||r|r|r|r|}
\hline
 \multirow{3}{*}{\shortstack[c]{data set}} & \multicolumn{1}{c|}{\multirow{3}{*}{\shortstack[c]{$\anz$}}} & \multicolumn{1}{c|}{\multirow{3}{*}{\shortstack[c]{$\f$}}} & \multirow{3}{*}{\shortstack[c]{number of data\\points in under-\\represented class}} & \multirow{3}{*}{\shortstack[c]{largest balanced\\sample possible}} \\
 & & & & \\
 & & & & \\ \hline \hline
Telescope\footnotemark & 19020 & 10 & 6688 & 13376 \\ \hline
HIGGS\footnotemark & 11000000 & 28 & 5170877 & 10341754 \\ \hline 
SUSY\footnotemark & 5000000 & 18 & 2287827 & 4575654 \\ \hline
YearPredictionMSD\footnotemark & 515345 & 90 & 206945 & 413890 \\ \hline
cod-rna\footnotemark & 488565 & 8 & 162855 & 209884\footnotemark \\ \hline
\end{tabular}
\captionsetup{justification=centering}
\captionof{table}{Benchmark data sets for numerical experiments}
\label{table:benchmark_data_sets}
\end{center}
\end{table}

\footnotetext[2]{\url{https://www.kaggle.com/brsdincer/telescope-spectrum-gamma-or-hadron}}
\footnotetext[3]{\url{https://archive.ics.uci.edu/ml/datasets/HIGGS}, Whiteson et al.~\cite{baldi2014searching}}
\footnotetext[4]{\url{https://archive.ics.uci.edu/ml/datasets/SUSY}, Whiteson et al.~\cite{baldi2014searching}}
\footnotetext[5]{\url{https://archive.ics.uci.edu/ml/datasets/yearpredictionmsd}, transform from multiclass into binary class by dividing into classes ``before 2000'' and ``as from 2000''}
\footnotetext[6]{\url{https://www.csie.ntu.edu.tw/~cjlin/libsvmtools/datasets/binary.html}, Uzilov et al.~\cite{uzilov2006detection}}


\footnotetext[7]{largest balanced sample possible $\neq$ 2 $\cdot$ number of data points in underrepresented class, since we need to respect the training and test set distribution}

In our experiments, we compare the performance of the proposed NFFT-based method with two state-of-the-art sklearn machine learning models, namely \linebreak \emph{sklearn.kernel\_ridge.KernelRidge} and \emph{sklearn.svm.SVC}. \rev{To the best of our knowledge both methods do not employ any tailored matrix routines such as sketching.} In our computations, we fix the MIS-threshold to $0.0$, which means we include all features, the FastAdjacency-setup to \emph{default} and the tolerance of convergence in the CG algorithm to 1e-03. \rev{At this stage we did not observe a dramatic increase in the number of CG steps and so far have not considered any preconditioning. This might become necessary if more challenging data sets and parameter choices are made.} The kernels are equally weighted and the data is balanced and z-score normalized. The results express mean values over 10 runs for the corresponding sample size, where the parameters are tuned using GridSearch for each and every run. For the considered classifiers, we choose the parameter-grids shown in Table~\ref{table:param_grids}.

\begin{table}[htp]
\small
\begin{center}
\begin{tabular}{|c||c|c|}
\hline
 classifier & kernel coefficient & regularization parameter
 \\ \hline \hline
 NFFTKernelRidge & $\sigma: [10^{-3}, 10^{-2}, 10^{-1}, 1, 10^{1}, 10^{2}, 10^{3}]$ & $\lambda: [1, 10^{1}, 10^{2}, 10^{3}]$ \\ \hline
 sklearn KRR & $\gamma: [10^{6}, 10^{4}, 10^{2}, 1, 10^{-2}, 10^{-4}, 10^{-6}]$ & $\alpha: [1, 10^{1}, 10^{2}, 10^{3}]$ \\ \hline
 sklearn SVC & $\gamma: [10^{6}, 10^{4}, 10^{2}, 1, 10^{-2}, 10^{-4}, 10^{-6}]$ & $C: [1, 10^{-1}, 10^{-2}, 10^{-3}]$ \\ \hline
\end{tabular}
\captionsetup{justification=centering}
\captionof{table}{Parameter-grids for the considered classifiers}
\label{table:param_grids}
\end{center}
\end{table}

The respective parameter-grids for the distinct classifiers are equivalent, as the relations {$\alpha = \frac{1}{C} = \lambda$} and {$\gamma = \frac{1}{\sigma^{2}}$} hold. For large samples, only one run is performed, since the standard deviation is very small, \rev{when that many} data points are included. Those samples are marked with $^*$ in the following tables. For all but the cod-rna data set, where we respect the train and test set distribution, the test set is obtained using random 50 percent of the sample. For the entire experiment, the train-test-split is 50:50. The objects of study are the accuracy and the runtimes for fitting and predicting within the learning methods.

In Table~\ref{table:accuracy_runtime_telescope}, the results for the Telescope Data Set are shown. We notice that the NFFT-approach's runtime starts at quite a high level, especially in comparison with sklearn SVC. But with increasing sample size, our NFFTKernelRidge classifier overtakes the other ones, with the accuracy being very satisfying as we even achieve 0.1 and 0.3 percent more for the maximal balanced sample, respectively. While we beat the sklearn classifiers not until a sample size somewhat greater than 10000 in fitting time, we outrun them in prediction time with 5000 data points already. With this, the presented classifier is 1.85 \rev{or} 1.65 times as fast as the sklearn classifiers for the largest balanced sample possible.

\begin{table}[htp]
\begin{center}
\resizebox{\textwidth}{!}{%
\begin{tabular}{|c||c|c|c||r|r|r||r|r|r||r|r|r|}
\hline
\multirow{3}{*}{\shortstack[c]{sample\\ size}} & \multicolumn{3}{c||}{\textbf{best accuracy}} & \multicolumn{3}{c||}{\textbf{mean runtime fit}} & \multicolumn{3}{c||}{\textbf{mean runtime predict}} & \multicolumn{3}{c|}{\textbf{mean total runtime}} \\
\cline{2-13}
& \multicolumn{1}{c|}{\multirow{2}{*}{\shortstack[c]{NFFT\\KRR}}} & \multicolumn{2}{c||}{sklearn} & \multicolumn{1}{c|}{\multirow{2}{*}{\shortstack[c]{NFFT\\KRR}}} & \multicolumn{2}{c||}{sklearn} & \multicolumn{1}{c|}{\multirow{2}{*}{\shortstack[c]{NFFT\\KRR}}} & \multicolumn{2}{c||}{sklearn} & \multicolumn{1}{c|}{\multirow{2}{*}{\shortstack[c]{NFFT\\KRR}}} & \multicolumn{2}{c|}{sklearn} \\
 & & KRR & SVC & & \multicolumn{1}{c|}{KRR} & \multicolumn{1}{c||}{SVC} & & \multicolumn{1}{c|}{KRR} & \multicolumn{1}{c||}{SVC} & & \multicolumn{1}{c|}{KRR} & \multicolumn{1}{c|}{SVC} \\ \hline \hline
100 & 69.6 & 74.2 & 57.0 & 0.1091 & 0.0026 & 0.0008 & 0.0121 & 0.0011 & 0.0003 & 0.1212 & 0.0036 & 0.0011 \\ \hline
500 & 80.4 & 78.1 & 76.6 & 0.1600 & 0.0075 & 0.0036 & 0.0168 & 0.0029 & 0.0021 & 0.1768 & 0.0104 & 0.0056 \\ \hline
1000 & 81.3 & 79.7 & 78.3 & 0.2104 & 0.0121 & 0.0125 & 0.0225 & 0.0089 & 0.0082 & 0.2328 & 0.0210 & 0.0207 \\ \hline
5000* & 83.2 & 82.0 & 81.4 & 0.4207 & 0.2086 & 0.1694 & 0.0376 & 0.1138 & 0.1300 & 0.4583 & 0.3224 & 0.2994 \\ \hline
10000* & 83.9 & 83.8 & 83.7 & 0.8128 & 0.8929 & 0.7027 & 0.0675 & 0.3991 & 0.5210 & 0.8803 & 1.2929 & 1.2237 \\ \hline
13376* & 83.9 & 83.8 & 83.6 & 1.2122 & 1.7200 & 1.2932 & 0.0978 & 0.7063 & 0.8661 & 1.3100 & 2.4264 & 2.1594 \\ \hline
\end{tabular}}
\captionsetup{justification=centering}
\captionof{table}{Telescope Data Set - Accuracy and runtime of GridSearch on various balanced samples of different size}
\label{table:accuracy_runtime_telescope}
\end{center}
\end{table}

Since the NFFT-based fast summation enables us to reduce the quadratic computational complexity, so that we nearly reach a linear scaling, this method is worth while even more for large data sets. Therefore, we want to consider the larger HIGGS Data Set next, see Table~\ref{table:accuracy_runtime_higgs}. As before, our method pays off with increasing sample size. Again, we overtake the competitors in prediction earlier than in fitting. But now, it takes larger samples to beat the sklearn classifiers. This is due to the HIGGS Data Set having 28 features, whereas the Telescope Data Set only had 10 features. And involving more features means having more windows, so that we must use more NFFTs for the approximation, which requires more time. 

A huge drawback of sklearn KRR is that it does not work for large-scale samples at all. For 25000 training data points the kernel repeatedly dies and when starting it for 50000 or more training data points, the computations break, since the program is ``unable to allocate 18.6 GiB for an array with shape (50000, 50000) and data type float64''. This is why some entries in the tables are missing. The same was experienced earlier, when computing kernel-vector products with the standard multiplication.
In contrast, the proposed NFFT-approach runs kernel-vector multiplications for samples of this size absolutely trouble-free and in perfectly reasonable time. Hence, our method enables working with large-scale data, where the state-of-the-art methods completely fail without adding additional sparsity.

In our experiment, the largest sample size, where all classifiers are working, is 25000 data points, i.\,e.\ 12500 training and test data points each. Here, the \linebreak NFFTKernelRidge classifier is 1.64 \rev{or} 1.48 times as fast as the sklearn classifiers in fitting. In predicting, the proposed learning model is even 15.12 \rev{or} 8.8 times as fast. However, for this sample size, our accuracy is 2.2 \rev{or} 1.7 percent beneath. But with increasing sample size, the NFFTKernelRidge classifier achieves to catch up on the accuracy, so that it reaches 0.1 percent more than sklearn SVC for 200000 data points. In fitting, the NFFT-approach is 9.44 and in predicting even 80.92 times as fast.

\begin{table}[htp]
\begin{center}
\resizebox{\textwidth}{!}{%
\begin{tabular}{|c||c|c|c||r|r|r||r|r|r||r|r|r|}
\hline
\multirow{3}{*}{\shortstack[c]{sample\\ size}} & \multicolumn{3}{c||}{\textbf{best accuracy}} & \multicolumn{3}{c||}{\textbf{mean runtime fit}} & \multicolumn{3}{c||}{\textbf{mean runtime predict}} & \multicolumn{3}{c|}{\textbf{mean total runtime}} \\
\cline{2-13}
& \multicolumn{1}{c|}{\multirow{2}{*}{\shortstack[c]{NFFT\\KRR}}} & \multicolumn{2}{c||}{sklearn} & \multicolumn{1}{c|}{\multirow{2}{*}{\shortstack[c]{NFFT\\KRR}}} & \multicolumn{2}{c||}{sklearn} & \multicolumn{1}{c|}{\multirow{2}{*}{\shortstack[c]{NFFT\\KRR}}} & \multicolumn{2}{c||}{sklearn} & \multicolumn{1}{c|}{\multirow{2}{*}{\shortstack[c]{NFFT\\KRR}}} & \multicolumn{2}{c|}{sklearn} \\
 & & KRR & SVC & & \multicolumn{1}{c|}{KRR} & \multicolumn{1}{c||}{SVC} & & \multicolumn{1}{c|}{KRR} & \multicolumn{1}{c||}{SVC} & & \multicolumn{1}{c|}{KRR} & \multicolumn{1}{c|}{SVC} \\ \hline \hline
100 & 55.8 & 54.4 & 40.4 & 0.2624 & 0.0020 & 0.0011 & 0.0344 & 0.0008 & 0.0006 & 0.2968 & 0.0028 & 0.0017 \\ \hline
500 & 58.8 & 55.7 & 54.3 & 0.3561 & 0.0048 & 0.0031 & 0.0392 & 0.0020 & 0.0020 & 0.3953 & 0.0068 & 0.0051 \\ \hline
1000 & 62.1 & 60.5 & 59.3 & 0.4325 & 0.0133 & 0.0107 & 0.0450 & 0.0040 & 0.0080 & 0.4776 & 0.0173 & 0.0187 \\ \hline
5000* & 63.3 & 62.9 & 62.0 & 1.0610 & 0.1914 & 0.2557 & 0.1056 & 0.1056 & 0.2084 & 1.1666 & 0.2970 & 0.4641 \\ \hline
10000* & 66.7 & 65.5 & 65.0 & 2.1512 & 0.8663 & 1.0339 & 0.1992 & 0.3673 & 0.7972 & 2.3504 & 1.2336 & 1.8312 \\ \hline
25000* & 64.8 & 67.0 & 66.5 & 6.3951 & 10.4802 & 9.4575 & 0.4987 & 7.5416 & 4.3869 & 6.8938 & 18.0218 & 13.8443 \\ \hline
50000* & 65.5 & \multicolumn{1}{c|}{-} & 67.3 & 14.4947 & \multicolumn{1}{c|}{-} & 56.6365 & 0.9397 & \multicolumn{1}{c|}{-} & 20.4737 & 15.4344 & \multicolumn{1}{c|}{-} & 77.1102 \\ \hline
100000* & 67.1 & \multicolumn{1}{c|}{-} & 67.9 & 34.6103 & \multicolumn{1}{c|}{-} & 197.9109 & 1.8479 & \multicolumn{1}{c|}{-} & 84.8191 & 36.4582 & \multicolumn{1}{c|}{-} & 282.7300 \\ \hline
200000* & 68.6 & \multicolumn{1}{c|}{-} & 68.5 & 105.0926 & \multicolumn{1}{c|}{-} & 991.7101 & 4.1666 & \multicolumn{1}{c|}{-} & 337.1600 & 109.2592 & \multicolumn{1}{c|}{-} & 1328.8701 \\ \hline
\end{tabular}}
\captionsetup{justification=centering}
\captionof{table}{HIGGS Data Set - Accuracy and runtime of GridSearch on various balanced samples of different size}
\label{table:accuracy_runtime_higgs}
\end{center}
\end{table}

To confirm our observations once again, we run this experiment on 3 more benchmark data sets. For the SUSY Data Set, see Table~\ref{table:accuracy_runtime_susy}, we trail the sklearn classifiers by 0.5 \rev{or} 0.3 percent in accuracy, but are 5.31 \rev{or} 4.78 times as fast in total runtime for a sample size of 50000. And when working with 200000 data points, the NFFT-based method is 0.4 percent worse than sklearn SVC in accuracy, but 16.7 times as fast.

\begin{table}[htp]
\begin{center}
\resizebox{\textwidth}{!}{%
\begin{tabular}{|c||c|c|c||r|r|r||r|r|r||r|r|r|}
\hline
\multirow{3}{*}{\shortstack[c]{sample\\ size}} & \multicolumn{3}{c||}{\textbf{best accuracy}} & \multicolumn{3}{c||}{\textbf{mean runtime fit}} & \multicolumn{3}{c||}{\textbf{mean runtime predict}} & \multicolumn{3}{c|}{\textbf{mean total runtime}} \\
\cline{2-13}
& \multicolumn{1}{c|}{\multirow{2}{*}{\shortstack[c]{NFFT\\KRR}}} & \multicolumn{2}{c||}{sklearn} & \multicolumn{1}{c|}{\multirow{2}{*}{\shortstack[c]{NFFT\\KRR}}} & \multicolumn{2}{c||}{sklearn} & \multicolumn{1}{c|}{\multirow{2}{*}{\shortstack[c]{NFFT\\KRR}}} & \multicolumn{2}{c||}{sklearn} & \multicolumn{1}{c|}{\multirow{2}{*}{\shortstack[c]{NFFT\\KRR}}} & \multicolumn{2}{c|}{sklearn} \\
 & & KRR & SVC & & \multicolumn{1}{c|}{KRR} & \multicolumn{1}{c||}{SVC} & & \multicolumn{1}{c|}{KRR} & \multicolumn{1}{c||}{SVC} & & \multicolumn{1}{c|}{KRR} & \multicolumn{1}{c|}{SVC} \\ \hline \hline
100 & 62.2 & 62.0 & 55.6 & 0.1218 & 0.0009 & 0.0005 & 0.0159 & 0.0004 & 0.0001 & 0.1377 & 0.0014 & 0.0006 \\ \hline
500 & 74.3 & 74.2 & 71.8 & 0.1754 & 0.0044 & 0.0028 & 0.0216 & 0.0029 & 0.0016 & 0.1969 & 0.0073 & 0.0044 \\ \hline
1000 & 76.8 & 75.5 & 73.6 & 0.2427 & 0.0114 & 0.0091 & 0.0307 & 0.0039 & 0.0058 & 0.2734 & 0.0153 & 0.0149 \\ \hline
5000* & 77.6 & 77.3 & 76.9 & 0.8570 & 0.1879 & 0.2033 & 0.0775 & 0.0939 & 0.1614 & 0.9345 & 0.2817 & 0.3647 \\ \hline
10000* & 78.4 & 78.4 & 78.0 & 1.5382 & 1.0586 & 0.8214 & 0.1356 & 0.3487 & 0.6078 & 1.6737 & 1.4073 & 1.4291 \\ \hline
50000* & 78.9 & 79.4 & 79.2 & 9.1113 & 42.7143 & 31.1025 & 0.6305 & 9.0420 & 15.5091 & 9.7417 & 51.7563 & 46.6116 \\ \hline
100000* & 78.7 & \multicolumn{1}{c|}{-} & 79.2 & 23.9190 & \multicolumn{1}{c|}{-} & 148.1214 & 1.2657 & \multicolumn{1}{c|}{-} & 62.3419 & 25.1847 & \multicolumn{1}{c|}{-} & 210.4633 \\ \hline
200000* & 78.9 & \multicolumn{1}{c|}{-} & 79.3 & 56.7895 & \multicolumn{1}{c|}{-} & 732.4476 & 2.3603 & \multicolumn{1}{c|}{-} & 255.1349 & 59.1497 & \multicolumn{1}{c|}{-} & 987.5824 \\ \hline
\end{tabular}}
\captionsetup{justification=centering}
\captionof{table}{SUSY Data Set - Accuracy and runtime of GridSearch on various balanced samples of different size}
\label{table:accuracy_runtime_susy}
\end{center}
\end{table}

For the YearPredictionMSD Data Set, see Table~\ref{table:accuracy_runtime_year}, the accuracy yielded by our NFFTKernelRidge classifier is 2.4 percent worse than sklearn SVC's accuracy, when 413890 data points are included. Even though using the NFFT-method makes the learning task 15.9 times faster, the forfeitures in accuracy are not negligible. By construction, the NFFT-approach only covers relations between the features being in the same window. The sklearn classifiers involve all relations, instead. It is easily conceivable, that the NFFT-method might lack one or another feature-connection, so that it cannot always get at the sklearn classifiers' accuracy. But we observe that the NFFTKernelRidge classifier obtains great runtimes even for data with 90 features, i.\,e.\ with 30 windows. It highly depends on the exact data set and classification task, whether it is worth taking 16 times as long for 2 percent more accuracy.

\begin{table}[htp]
\begin{center}
\resizebox{\textwidth}{!}{%
\begin{tabular}{|c||c|c|c||r|r|r||r|r|r||r|r|r|}
\hline
\multirow{3}{*}{\shortstack[c]{sample\\ size}} & \multicolumn{3}{c||}{\textbf{best accuracy}} & \multicolumn{3}{c||}{\textbf{mean runtime fit}} & \multicolumn{3}{c||}{\textbf{mean runtime predict}} & \multicolumn{3}{c|}{\textbf{mean total runtime}} \\
\cline{2-13}
& \multicolumn{1}{c|}{\multirow{2}{*}{\shortstack[c]{NFFT\\KRR}}} & \multicolumn{2}{c||}{sklearn} & \multicolumn{1}{c|}{\multirow{2}{*}{\shortstack[c]{NFFT\\KRR}}} & \multicolumn{2}{c||}{sklearn} & \multicolumn{1}{c|}{\multirow{2}{*}{\shortstack[c]{NFFT\\KRR}}} & \multicolumn{2}{c||}{sklearn} & \multicolumn{1}{c|}{\multirow{2}{*}{\shortstack[c]{NFFT\\KRR}}} & \multicolumn{2}{c|}{sklearn} \\
 & & KRR & SVC & & \multicolumn{1}{c|}{KRR} & \multicolumn{1}{c||}{SVC} & & \multicolumn{1}{c|}{KRR} & \multicolumn{1}{c||}{SVC} & & \multicolumn{1}{c|}{KRR} & \multicolumn{1}{c|}{SVC} \\ \hline \hline
100 & 46.0 & 50.8 & 42.6 & 0.6578 & 0.0011 & 0.0007 & 0.0899 & 0.0005 & 0.0006 & 0.7476 & 0.0016 & 0.0013 \\ \hline
500 & 61.4 & 64.2 & 64.2 & 0.8981 & 0.0051 & 0.0064 & 0.1122 & 0.0022 & 0.0054 & 1.0103 & 0.0074 & 0.0118 \\ \hline
1000 & 66.0 & 69.0 & 68.2 & 1.1817 & 0.0107 & 0.0218 & 0.1405 & 0.0066 & 0.0196 & 1.3221 & 0.0173 & 0.0413 \\ \hline
5000* & 70.3 & 71.7 & 71.7 & 3.2502 & 0.1861 & 0.5164 & 0.3882 & 0.0966 & 0.4705 & 3.6384 & 0.2827 & 0.9869 \\ \hline
10000* & 72.3 & 73.4 & 73.1 & 6.7636 & 0.8400 & 2.1104 & 0.7016 & 0.3625 & 1.8575 & 7.4651 & 1.2025 & 3.9679 \\ \hline
50000* & 73.4 & 74.8 & 74.8 & 53.3774 & 43.2971 & 122.4035 & 3.9307 & 9.1279 & 50.3554 & 57.3081 & 52.4250 & 172.7589 \\ \hline
100000* & 74.2 & \multicolumn{1}{c|}{-} & 75.8 & 102.0414 & \multicolumn{1}{c|}{-} & 426.8239 & 6.1424 & \multicolumn{1}{c|}{-} & 201.7097 & 108.1838 & \multicolumn{1}{c|}{-} & 628.5336 \\ \hline
200000* & 74.4 & \multicolumn{1}{c|}{-} & 76.4 & 245.1812 & \multicolumn{1}{c|}{-} & 2006.6694 & 12.3315 & \multicolumn{1}{c|}{-} & 785.3145 & 257.5127 & \multicolumn{1}{c|}{-} & 2791.9839 \\ \hline
413890* & 74.3 & \multicolumn{1}{c|}{-} & 76.7 & 669.4028 & \multicolumn{1}{c|}{-} & 7696.3063 & 24.4887 & \multicolumn{1}{c|}{-} & 3335.6310 & 693.8915 & \multicolumn{1}{c|}{-} & 11031.9373 \\ \hline
\end{tabular}}
\captionsetup{justification=centering}
\captionof{table}{YearPredictionMSD Data Set - Accuracy and runtime of GridSearch on various balanced samples of different size}
\label{table:accuracy_runtime_year}
\end{center}
\end{table}

For the cod-rna Data Set, see Table~\ref{table:accuracy_runtime_cod-rna}, NFFTKernelRidge's accuracy is only 0.5 percent worse than sklearn SVC's for 209884 data points, the largest balanced sample possible. This slight loss will be acceptable in most cases, when being 19.77 times as fast in return.

\begin{table}[htp]
\begin{center}
\resizebox{\textwidth}{!}{%
\begin{tabular}{|c||c|c|c||r|r|r||r|r|r||r|r|r|}
\hline
\multirow{3}{*}{\shortstack[c]{sample\\ size}} & \multicolumn{3}{c||}{\textbf{best accuracy}} & \multicolumn{3}{c||}{\textbf{mean runtime fit}} & \multicolumn{3}{c||}{\textbf{mean runtime predict}} & \multicolumn{3}{c|}{\textbf{mean total runtime}} \\
\cline{2-13}
& \multicolumn{1}{c|}{\multirow{2}{*}{\shortstack[c]{NFFT\\KRR}}} & \multicolumn{2}{c||}{sklearn} & \multicolumn{1}{c|}{\multirow{2}{*}{\shortstack[c]{NFFT\\KRR}}} & \multicolumn{2}{c||}{sklearn} & \multicolumn{1}{c|}{\multirow{2}{*}{\shortstack[c]{NFFT\\KRR}}} & \multicolumn{2}{c||}{sklearn} & \multicolumn{1}{c|}{\multirow{2}{*}{\shortstack[c]{NFFT\\KRR}}} & \multicolumn{2}{c|}{sklearn} \\
 & & KRR & SVC & & \multicolumn{1}{c|}{KRR} & \multicolumn{1}{c||}{SVC} & & \multicolumn{1}{c|}{KRR} & \multicolumn{1}{c||}{SVC} & & \multicolumn{1}{c|}{KRR} & \multicolumn{1}{c|}{SVC} \\ \hline \hline
100 & 78.0 & 77.2 & 72.2 & 0.0459 & 0.0010 & 0.0005 & 0.0066 & 0.0006 & 0.0002 & 0.0524 & 0.0016 & 0.0007 \\ \hline
500 & 89.2 & 92.7 & 89.6 & 0.0671 & 0.0039 & 0.0021 & 0.0074 & 0.0023 & 0.0012 & 0.0745 & 0.0063 & 0.0034 \\ \hline
1000 & 92.5 & 94.7 & 93.7 & 0.0882 & 0.0110 & 0.0071 & 0.0100 & 0.0037 & 0.0044 & 0.0982 & 0.0147 & 0.0116 \\ \hline
5000* & 95.4 & 95.7 & 95.6 & 0.3046 & 0.1788 & 0.1564 & 0.0274 & 0.0952 & 0.1031 & 0.3320 & 0.2740 & 0.2595 \\ \hline
10000* & 95.7 & 95.8 & 95.9 & 0.5864 & 0.8309 & 0.6044 & 0.0472 & 0.3723 & 0.4101 & 0.6336 & 1.2032 & 1.0145 \\ \hline
50000* & 95.7 & 95.9 & 96.0 & 3.9697 & 43.4380 & 17.9735 & 0.2376 & 9.5084 & 9.8435 & 4.2073 & 52.9464 & 27.8170 \\ \hline
100000* & 95.9 & \multicolumn{1}{c|}{-} & 96.2 & 10.1743 & \multicolumn{1}{c|}{-} & 73.8995 & 0.4686 & \multicolumn{1}{c|}{-} & 39.0615 & 10.6430 & \multicolumn{1}{c|}{-} & 112.9610 \\ \hline
209884* & 96.0 & \multicolumn{1}{c|}{-} & 96.5 & 28.9638 & \multicolumn{1}{c|}{-} & 422.5391 & 1.0058 & \multicolumn{1}{c|}{-} & 170.0743 & 29.9696 & \multicolumn{1}{c|}{-} & 592.6134 \\ \hline
\end{tabular}}
\captionsetup{justification=centering}
\captionof{table}{cod-rna Data Set - Accuracy and runtime of GridSearch on various balanced samples of different size}
\label{table:accuracy_runtime_cod-rna}
\end{center}
\end{table}

\begin{figure}[htp]
\begin{center}
\includegraphics[scale=0.25]{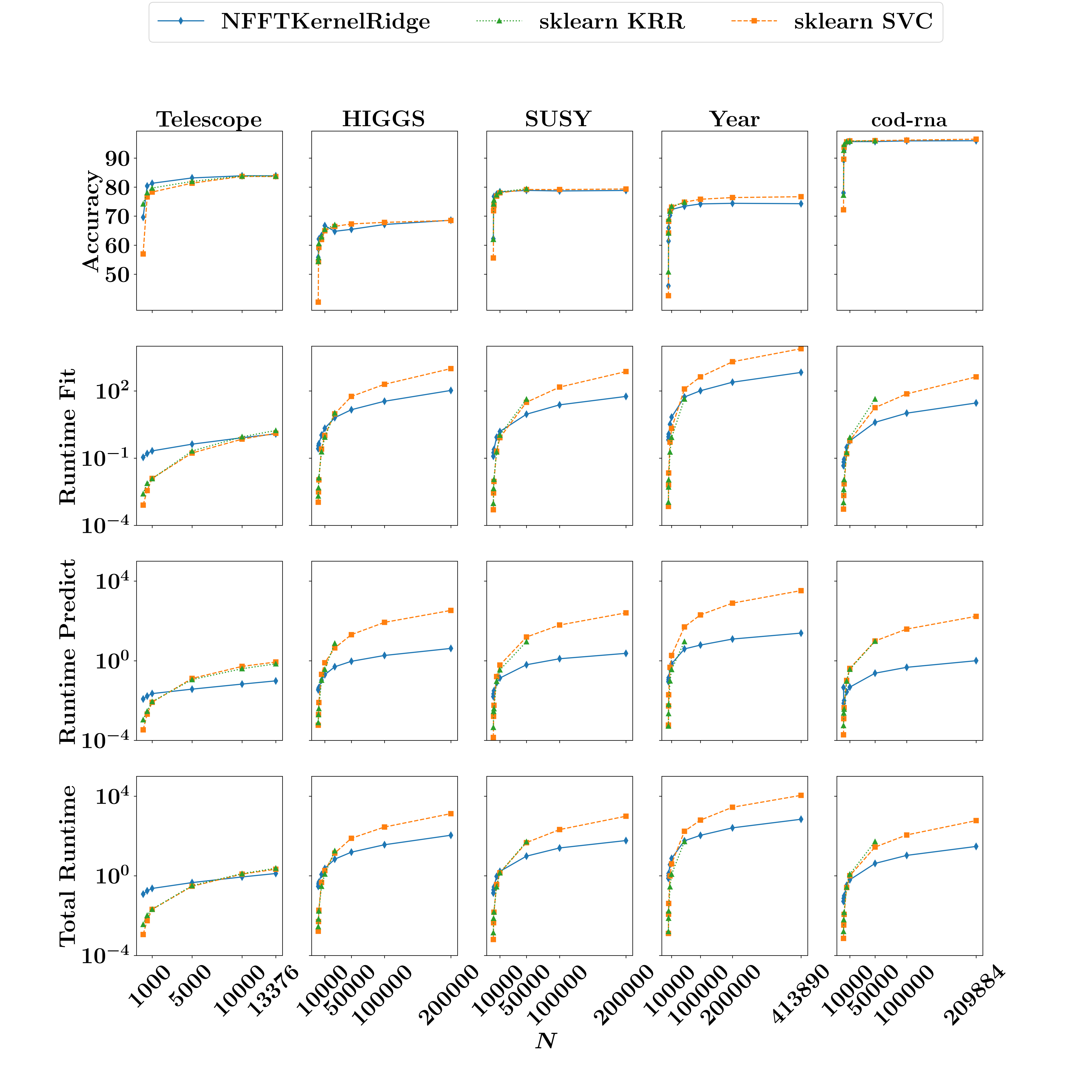}
\caption{Accuracy and runtime of GridSearch on various balanced samples of different size}
\label{fig:accuracy_and_runtime_all_data_sets}
\end{center}
\end{figure}

In Figure~\ref{fig:accuracy_and_runtime_all_data_sets}, the results from Tables~\ref{table:accuracy_runtime_telescope} to \ref{table:accuracy_runtime_cod-rna} are visualized. While the graphs representing the accuracy almost align for all classifiers on all data sets considered, the graphs showing the runtimes give a totally different picture. Again, it is clearly visible that the NFFTKernelRidge classifier avails to nothing for small data sets. But for large-scale data its runtime beats the competitive classifiers by several magnitudes. This applies for prediction in particular. Notice that the runtimes highly depend on the number of features, i.\,e.\ windows, involved and also on the choice of parameters. 

\section{Summary and Outlook}
We have studied the applicability of NFFT-based fast summation on kernel methods with a high-dimensional feature space. The ANOVA kernel has proved to be a viable tool to group the features into smaller pieces that are then amenable to the NFFT-based summation technique. We embedded this approach into a kernel ridge regression scheme that requires matrix-vector products as part of the CG algorithm to solve the linear system. \rev{We have illustrated} the competitiveness of our method on several large-scale examples.

As a future topic we would like to examine the relationship between our method and the algorithms proposed by Potts and Schmischke~\cite{PoSchmi19,PoSchmi20}, which allow for a sensitivity analysis by studying truncated ANOVA decompositions of functions with low-dimensional structures in terms of the Fourier coefficients.

Additionally, we believe our scheme can be embedded into \rev{methods that rely on kernel matrices or even graph Laplacians. Examples are support vector machines or graph neural networks but also other loss functions that rely on kernel formulations and can then benefit from having the fast matrix-vector product available.}

\section*{Acknowledgment}
The authors gratefully acknowledge their support by the \linebreak BMBF grant 01|S20053A (project SA$\ell$E).
We thank our colleagues in the project SA$\ell$E for valuable discussions on the contents of this paper.










\begin{thebibliography}{10}

\bibitem{adeyemo2019effects}
A.~Adeyemo, H.~Wimmer, and L.~M. Powell.
\newblock {Effects of Normalization Techniques on Logistic Regression in Data
  Science}.
\newblock {\em Journal of Information Systems Applied Research}, 12(2):37,
  2019.

\bibitem{alfke2018nfft}
D.~Alfke, D.~Potts, M.~Stoll, and T.~Volkmer.
\newblock {NFFT Meets Krylov Methods: Fast Matrix-Vector Products for the Graph
  Laplacian of Fully Connected Networks}.
\newblock {\em Frontiers in Applied Mathematics and Statistics}, 4:61, 2018.

\bibitem{avron2017faster}
H.~Avron, K.~L. Clarkson, and D.~P. Woodruff.
\newblock {Faster Kernel Ridge Regression Using Sketching and Preconditioning}.
\newblock {\em SIAM J. Matrix Anal. Appl.}, 38(4):1116--1138, 2017.

\bibitem{baldi2014searching}
P.~Baldi, P.~Sadowski, and D.~Whiteson.
\newblock {Searching for exotic particles in high-energy physics with deep
  learning}.
\newblock {\em Nature Communications}, 5(1):1--9, 2014.

\bibitem{battiti1994using}
R.~Battiti.
\newblock {Using mutual information for selecting features in supervised neural
  net learning}.
\newblock {\em IEEE Transactions on Neural Networks}, 5(4):537--550, 1994.

\bibitem{bey95}
G.~Beylkin.
\newblock {On the Fast Fourier Transform of Functions with Singularities}.
\newblock {\em Applied Computational Harmonic Analysis}, 2:363--381, 1995.

\bibitem{bishop2006pattern}
C.~M. Bishop and N.~M. Nasrabadi.
\newblock {\em Pattern Recognition and Machine Learning}.
\newblock Springer, 2006.

\bibitem{donders2006gentle}
A.~R.~T. Donders, G.~J. Van Der~Heijden, T.~Stijnen, and K.~G. Moons.
\newblock {A gentle introduction to imputation of missing values}.
\newblock {\em Journal of Clinical Epidemiology}, 59(10):1087--1091, 2006.

\bibitem{DuSc}
A.~J.~W. Duijndam and M.~A. Schonewille.
\newblock {Nonuniform fast Fourier transform}.
\newblock {\em GEOPHYSICS}, 64:539--551, 1999.

\bibitem{duro93}
A.~Dutt and V.~Rokhlin.
\newblock {Fast Fourier Transforms for Nonequispaced Data}.
\newblock {\em SIAM J. Sci. Comput.}, 14:1368--1393, 1993.

\bibitem{FeSt03}
M.~Fenn and G.~Steidl.
\newblock {Fast NFFT Based Summation of Radial Functions}.
\newblock {\em Sampling Theory in Signal and Image Processing}, 3:1--28, 2004.

\bibitem{golub2013matrix}
G.~H. Golub and C.~F. Van~Loan.
\newblock {\em {Matrix Computations}}.
\newblock JHU Press, 2013.

\bibitem{gonen2011multiple}
M.~G{\"o}nen and E.~Alpayd{\i}n.
\newblock {Multiple kernel learning algorithms}.
\newblock {\em J. Mach. Learn. Res.}, 12:2211--2268, 2011.

\bibitem{hestenes1952methods}
M.~R. Hestenes, E.~Stiefel, et~al.
\newblock {Methods of Conjugate Gradients for Solving Linear Systems}.
\newblock {\em Journal of Research of the National Bureau of Standards},
  49(6):409, 1952.

\bibitem{hofmann2008kernel}
T.~Hofmann, B.~Sch{\"o}lkopf, and A.~J. Smola.
\newblock {Kernel methods in machine learning}.
\newblock {\em Ann. Statist.}, 36(3):1171--1220, 2008.

\bibitem{KeKuPo09}
J.~Keiner, S.~Kunis, and D.~Potts.
\newblock {Using NFFT3- A Software Library for Various Nonequispaced Fast
  Fourier Transforms}.
\newblock {\em ACM Trans. Math. Software}, 36:Article 19, 1--30, 2009.

\bibitem{kipf2016semi}
T.~N. Kipf and M.~Welling.
\newblock {Semi-Supervised Classification with Graph Convolutional Networks}.
\newblock {\em arXiv preprint arXiv:1609.02907}, 2016.

\bibitem{ASKIT}
W.~March, B.~Xiao, and G.~Biros.
\newblock {ASKIT: Approximate Skeletonization Kernel-Independent Treecode in
  High Dimensions}.
\newblock {\em SIAM J. Sci. Comput.}, 37, 10 2014.

\bibitem{marques2013suitability}
A.~I. Marqu{\'e}s, V.~Garc{\'\i}a, and J.~S. S{\'a}nchez.
\newblock {On the suitability of resampling techniques for the class imbalance
  problem in credit scoring}.
\newblock {\em Journal of the Operational Research Society}, 64(7):1060--1070,
  2013.

\bibitem{morariu2008automatic}
V.~I. Morariu, B.~V. Srinivasan, V.~C. Raykar, R.~Duraiswami, and L.~S. Davis.
\newblock {Automatic online tuning for fast Gaussian summation}.
\newblock {\em Advances in Neural Information Processing Systems}, 21, 2008.

\bibitem{patro2015normalization}
S.~Patro and K.~K. Sahu.
\newblock Normalization: A preprocessing stage.
\newblock {\em arXiv preprint arXiv:1503.06462}, 2015.

\bibitem{PoSchmi19}
D.~Potts and M.~Schmischke.
\newblock {Approximations of High-Dimensional Periodic Functions with
  Fourier-Based Methods}.
\newblock {\em SIAM J. Numer. Anal. 59}, pages 2393--2429, 2021.
\newblock arXiv:1907.11412 [math.NA].

\bibitem{PoSchmi20}
D.~Potts and M.~Schmischke.
\newblock {Learning multivariate functions with low-dimensional structures
  using polynomial bases}.
\newblock {\em J. Comput. Appl. Math.}, 403:113821, 2022.

\bibitem{post02}
D.~Potts and G.~Steidl.
\newblock {Fast Summation at Nonequispaced Knots by NFFT}.
\newblock {\em SIAM J. Sci. Comput.}, 24:2013--2037, 2003.

\bibitem{postni04}
D.~Potts, G.~Steidl, and A.~Nieslony.
\newblock {Fast convolution with radial kernels at nonequispaced knots}.
\newblock {\em Numer. Math.}, 98:329--351, 2004.

\bibitem{rasmussen2003gaussian}
C.~E. Rasmussen.
\newblock {Gaussian Processes in Machine Learning}.
\newblock In {\em Summer School on Machine Learning}, pages 63--71. Springer,
  2003.

\bibitem{raykar2007fast}
V.~C. Raykar and R.~Duraiswami.
\newblock {Fast large scale Gaussian process regression using approximate
  matrix-vector products}.
\newblock In {\em Learning workshop}, 2007.

\bibitem{saad2003iterative}
Y.~Saad.
\newblock {\em {Iterative methods for sparse linear systems}}.
\newblock SIAM, 2003.

\bibitem{sarlecomp}
W.~Sarle.
\newblock {comp. ai. neural-nets FAQ, Part 2 of 7: Learning}.
\newblock \url{http://www. faqs. org/faqs/ai-faq/neural-nets/part2}, 1997
  (accessed: 22~February 2021).

\bibitem{scholkopf2002learning}
B.~Sch{\"o}lkopf and A.~J. Smola.
\newblock {\em {Learning with Kernels: Support Vector Machines, Regularization,
  Optimization, and Beyond}}.
\newblock MIT Press, 2002.

\bibitem{shawe2004kernel}
J.~Shawe-Taylor, N.~Cristianini, et~al.
\newblock {\em {Kernel Methods for Pattern Analysis}}.
\newblock Cambridge University Press, 2004.

\bibitem{st97}
G.~Steidl.
\newblock {A note on fast Fourier transforms for nonequispaced grids}.
\newblock {\em Adv. Comput. Math.}, 9:337--353, 1998.

\bibitem{stoll2020literature}
M.~Stoll.
\newblock {A literature survey of matrix methods for data science}.
\newblock {\em GAMM-Mitt.}, 43(3):e202000013, 2020.

\bibitem{tanabe2008simple}
H.~Tanabe, T.~B. Ho, C.~H. Nguyen, and S.~Kawasaki.
\newblock {Simple but effective methods for combining kernels in computational
  biology}.
\newblock In {\em 2008 IEEE International Conference on Research, Innovation
  and Vision for the Future in Computing and Communication Technologies}, pages
  71--78. IEEE, 2008.

\bibitem{uzilov2006detection}
A.~V. Uzilov, J.~M. Keegan, and D.~H. Mathews.
\newblock {Detection of non-coding RNAs on the basis of predicted secondary
  structure formation free energy change}.
\newblock {\em BMC Bioinformatics}, 7(1):1--30, 2006.

\bibitem{wilson2016deep}
A.~G. Wilson, Z.~Hu, R.~Salakhutdinov, and E.~P. Xing.
\newblock {Deep Kernel Learning}.
\newblock In {\em Artificial Intelligence and Statistics}, pages 370--378.
  Proc. Mach. Learn. Res. (PMLR), 2016.

\bibitem{INV-ASKIT}
C.~Yu, W.~March, B.~Xiao, and G.~Biros.
\newblock {INV-ASKIT: A Parallel Fast Direct Solver for Kernel Matrices}.
\newblock {\em 2016 IEEE International Parallel and Distributed Processing
  Symposium}, pages 161--171, 2016.

\bibitem{zheng2018feature}
A.~Zheng and A.~Casari.
\newblock {\em {Feature Engineering for Machine Learning: Principles and
  Techniques for Data Scientists}}.
\newblock O'Reilly Media, Inc., 2018.

\end{thebibliography}

\medskip
Received xxxx 20xx; revised xxxx 20xx; early access xxxx 20xx.
\medskip

\end{document}